\documentclass[letterpaper, 10 pt, journal, table]{IEEEtran}  

\IEEEoverridecommandlockouts                              

\usepackage{graphics} 
\usepackage{float}
\usepackage{cite}
\usepackage{adjustbox}
\usepackage{mathptmx} 
\usepackage{stfloats}
\usepackage{amsmath} 
\usepackage{amssymb}  
\usepackage{todonotes}
\usepackage{multirow}
\usepackage{algorithm}
\usepackage{algpseudocode}
\usepackage{subcaption}
\usepackage{color,soul}
\makeatother
\usepackage[colorlinks, linkcolor=green, citecolor=blue]{hyperref}

\def\BibTeX{{\rm B\kern-.05em{\sc i\kern-.025em b}\kern-.08em
  T\kern-.1667em\lower.7ex\hbox{E}\kern-.125emX}}
  
\title{\LARGE \bf
USPilot: An Embodied Robotic Assistant Ultrasound System with Large Language Model Enhanced Graph Planner
}

\author{ Mingcong Chen$^{1,3}$ Siqi Fan$^{2,3}$ Guanglin Cao$^{3,4}$ Yun-hui Liu$^{2}$ Hongbin Liu$^{3,4,5}$
\thanks{*This work was supported by InnoHK and in part of the HK RGC AoE under AoE/E-407/24-N.}
\thanks{*This study obtained ethical approval from the Institute of Automation, Chinese Academy of Sciences local ethics committee (study title: Ultrasound Robot Motion Perception and Control Based on Multimodal Sensing, study reference: IA21-2502-020302).}
\thanks{$^{1}$ Mingcong Chen is with the Department of Biomedical Engineering, City University of Hong Kong, Hong Kong SAR, $^{2}$ Siqi Fan and Yun-hui Liu are with the Department of Mechanical and Automation Engineering, Chinese University of Hong Kong, HKSAR, $^{3}$ Mingcong Chen, Siqi Fan, Guanglin Cao and Hongbin Liu are with the Centre for Artificial Intelligence and Robotics Hong Kong Institute of Science \& Innovation, Chinese Academy of Sciences, HKSAR, $^{4}$ Guanglin Cao and Hongbin Liu are with the Institute of Automation, Chinese Academy of Sciences, China, $^{5}$Hongbin Liu is with Department of Surgical and Interventional Engineering, King’s College London, UK.}%
\thanks{Mingcong Chen and Siqi Fan are co-first authors.}
\thanks{Correspondence: Hongbin Liu \tt{liuhongbin@ia.ac.cn}}
}

\begin{document}

\maketitle
\thispagestyle{empty}
\pagestyle{empty}

\begin{abstract}

In the era of Large Language Models (LLMs), embodied artificial intelligence presents transformative opportunities for robotic manipulation tasks. Ultrasound imaging, a widely used and cost-effective medical diagnostic procedure, faces challenges due to the global shortage of professional sonographers. To address this issue, we propose USPilot, an embodied robotic assistant ultrasound system powered by an LLM-based framework to enable autonomous ultrasound acquisition. USPilot is designed to function as a virtual sonographer, capable of responding to patients' ultrasound-related queries and performing ultrasound scans based on user intent. By fine-tuning the LLM, USPilot demonstrates a deep understanding of ultrasound-specific questions and tasks. Furthermore, USPilot incorporates an LLM-enhanced Graph Neural Network (GNN) to manage ultrasound robotic APIs and serve as a task planner. Experimental results show that the LLM-enhanced GNN achieves unprecedented accuracy in task planning on public datasets with an accuracy of 78.64\%, 60.8\% and 59.6\%. Additionally, the system demonstrates significant potential in autonomously understanding and executing ultrasound procedures with a successful demonstration of our physical setup with the robot. These advancements bring us closer to achieving autonomous and potentially unmanned robotic ultrasound systems, addressing critical resource gaps in medical imaging.

\end{abstract}

\section{Introduction}
\label{sec:intro}
Ultrasound is widely used in medical imaging. It provides real-time images of internal organs, blood vessels, and tissues without exposing patients to ionizing radiation. However, hospital sonographers face significant challenges due to the increasing demand for ultrasound services and the rapid pace of technological advancements. There are over 64,751 sonographers employed in the United States, but the number will need to increase by 27,600, according to a prediction by the Bureau of Labor Statistics\cite{murphey2017work}. In addition, sonographers commonly experience physical strain from their work. Repetitive motions often lead to joint discomfort, particularly in the wrists, upper back, and shoulders\cite{harrison2015work}. To mitigate this issue, it is imperative to consider the implementation of automation, particularly in light of the global shortage of ultrasound imaging professionals and the increasing demand for ultrasound-based diagnostic imaging.

The introduction of automated diagnostic and intraoperative ultrasound systems, which are either guided or assisted by robotic technology, represents a promising solution to the challenges previously outlined. 
There are currently several robotic assistant ultrasound scanning systems for specific tasks to enable more consistent digital images without sonographers,\cite{bi2024machine,jiang2024intelligent} such as reinforcement learning\cite{li2021autonomous,ning2021autonomic}, imitation learning\cite{men2022multimodal, deng2021learning}, reward learning\cite{jiang2024intelligent} and predefined tasks\cite{chen2023fully}. Specifically, Raina et al. propose a Bayesian optimization framework guided by an expert‑modeled “quality map” to direct probe movements\cite{raina2023robotic}. Jiang et al. use a disentangled rewarding learning method to capture the ``language of sonography'' through self-supervised pairwise image comparisons\cite{jiang2024intelligent}. They can achieve high accuracy, but these systems still require extensive domain knowledge or reward engineering, requiring doctors or engineers to select the specific functions individually. The human-robot interaction methods in these systems may also increase surgeons’ workload or distract them from their current tasks, which do not align with their operating habits and far away from future intelligent robotic sonographers\cite{jiang2023robotic}.

\begin{figure}[t]
      \centering
    \includegraphics[width=0.8\linewidth]{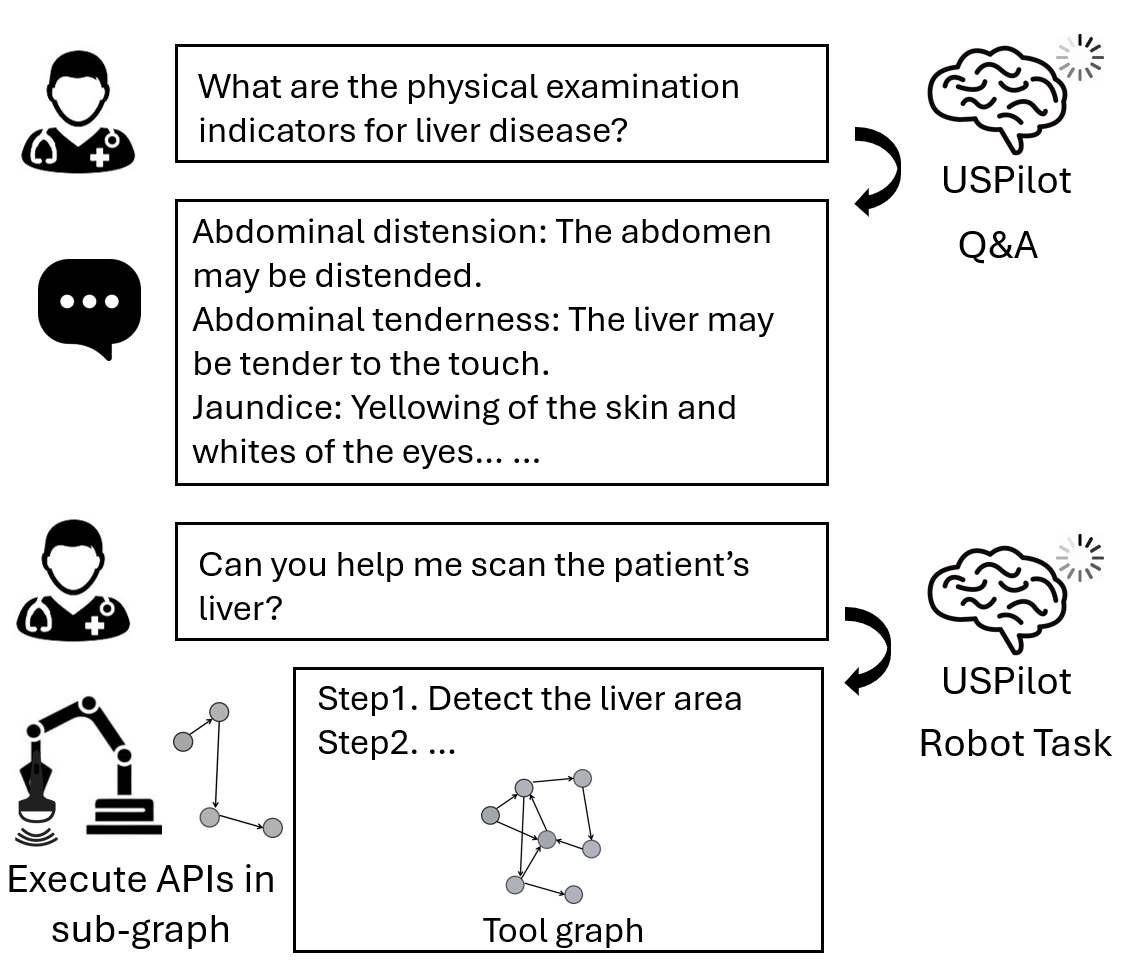}
      \caption{The overview of USPilot: Answer the sonographer's normal medical questions or perform an automated ultrasound scan.}
      \label{workflow}
\end{figure}

A subsequent surge in embodied artificial intelligence has been observed with the rise of large language models (LLMs) owing to their powerful associative capability and extensive prior knowledge. Some researchers have utilized its ability of understanding human text for task planning and reasoning, such as Octopus \cite{yang2023octopus} 
and TAPA \cite{wu2023embodied}. 
Other works go further on robot manipulation, such as \cite{brohan2023can, driess2023palm, huang2023voxposer} 
finetune the LLM on manipulating data together with pre-defined vision feedback. However, such low level robot control will easily face the illusion, which is unacceptable in medical scenarios.

In this paper, we present USPilot, a comprehensive framework designed for autonomous ultrasound acquisition. From the observation that when a human manipulates a device, they have an exact workflow in mind. The workflow is fixed due to the device being fixed, but it has various subpaths for different specific tasks. Hence, we defined the robotic ultrasound workflow as a graph and utilized the capability of LLMs to select a subgraph (i.e, a subpath as the human) based on the semantic information from the given task.
Following this basic idea, the framework incorporates a semantic router powered by LLMs that interprets user queries related to ultrasound procedures and translates natural language for robotic ultrasound system operations. A key innovation of our framework is the implementation of an LLM-enhanced Graph Neural Network (GNN) for robotic tool selection and planning. 
By leveraging adapter-based ultrasound knowledge integration, USPilot demonstrates successful autonomous ultrasound image acquisition on a physical robotic platform, marking a significant step toward fully autonomous robotic ultrasound systems. This advancement suggests promising possibilities for unmanned medical imaging applications.


\begin{figure*}[t]
      \centering
    \includegraphics[width=0.81\linewidth]{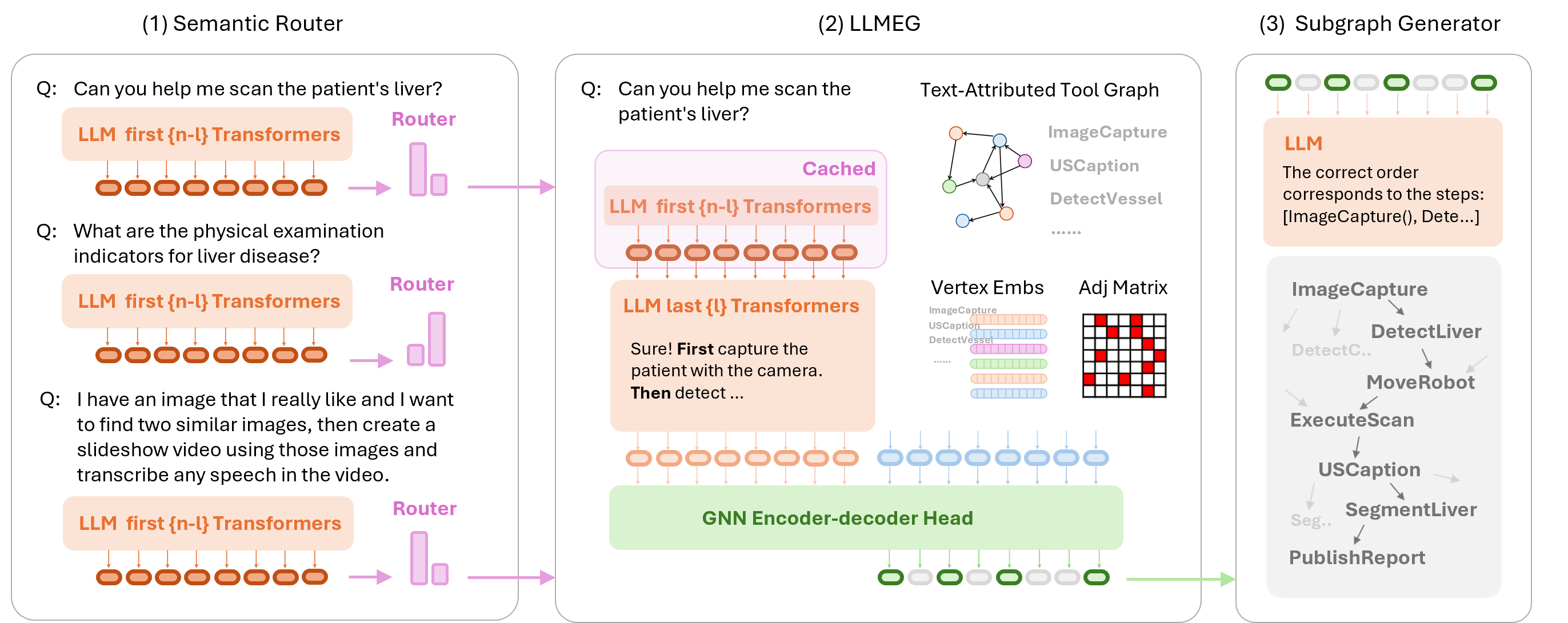}
      \caption{The structure of USPilot: (1) The semantic router recognizes the user’s intent. (2) If the intent is an executable task, LLMEG is invoked using the cached, unadapted Transformer to select potential APIs. (3) The LLM-based subgraph generator reorders the selected APIs into a directed graph.}
      \label{overview}
\end{figure*}
\vspace{-5pt}

\section{{Methodology}}

\subsection{Problem formulation}

In this work, the system adapts to the user's instruction types to perform ultrasound robot manipulation or answer ultrasound-related questions. During usage, the LLM-based planner receives a text instruction $I$ and a task description graph $G$. The proposed system can generate a text answer $A$ or a serial executable robotic command $G^*$ based on the user intent perception policy $\pi$. The answering process can be formulated as $A\bigvee G^*=\pi(I,G)$.

To build this task description graph $G$, note that a robotic control system naturally consists of low‑level functions, which can be exposed as related application programming interfaces (APIs). These executable APIs can form a graph where each vertex $v\in V$ corresponds to an API in the robotic system and each edge $(x,y)\in E$ indicates the dependency between two APIs. The robotic ultrasound manipulation can then be formulated as a task-planning problem on this fixed tool graph. However, unlike the structured graph, text instructions are in natural language, which is ambiguous and lies in another latent space. LLMs can perform as a perfect bridge as far as the API description text $t\in T$ is attached to each vertex to provide a linguistic description (eg. \textit{LiverSegmentation.Segments the liver from the ultrasound image.}). Thus, the manipulation problem based on $I$ turns into a subgraph extraction on the predefined TaG graph $G=(V,E,T)$ according to textual information. In general speaking, the graph-based task planning aims to identify an optimal subgraph $G^*$ that can complete the given task instruction $I$.

\begin{figure*}[t]
      \centering
    \includegraphics[width=0.9\linewidth]{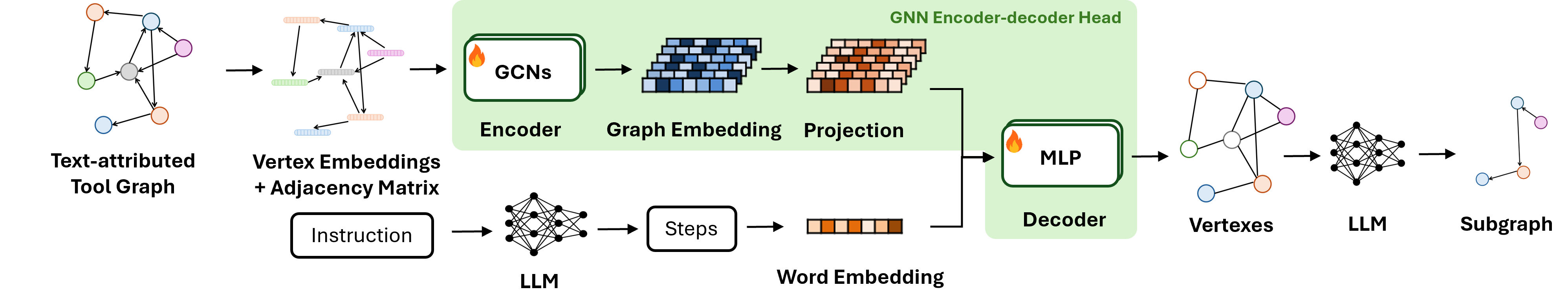}
      \caption{The structure of LLMEG: Select a toolchain $G^*$ from $G$ based on the textual information provided by $I$.}
      \label{llmeg}
\end{figure*}
\vspace{-5pt}

\subsection{Structure overview}
The overview of USPilot is depicted in Fig. \ref{overview}. The semantic router is a dynamic routing mechanism connected to the LLM transformer structure. It enables user intent recognition, which helps the system load the appropriate knowledge or graphs. Related subject knowledge is trained into the LLM through an adapter structure, allowing the model to better respond to user intent by adjusting the gating factor of the adapter layers. 

If the user intent involves executable tasks, the router switches to the Large Language Model Enhanced Graph (LLMEG) as Fig \ref{llmeg}. The LLMEG comprises an LLM-enhanced Graph Neural Network (GNN) head and a subgraph generator. The LLM-enhanced GNN head selects APIs by sending LLM-generated word embeddings and GNN-generated graph embeddings to the decoder. The LLM-based subgraph generator defines the path for API execution. When the user intent involves a general medical question, the router switches the inference path to a pre-trained ultrasound-related question-answer adapter to fulfill the medical inquiry requirements. The semantic router, adapters, and LLMEG are all hot-plugging in our USPilot, which makes it suitable for separate training, upgrading, and adapting.

\subsection{Large language model enhanced graph}
LLMs possess a remarkable ability to interpret user requests and decompose complex commands into simpler subtasks. This task decomposition capability is leveraged using the LLM to generate subtasks from input instructions, which are then converted into word embeddings, denoted as $e_w$. 
To inherit the API description text into each vertex, inspired by LLaMA2 technical report\cite{touvron2023llama}
, it is suggested that the latent space of LLM is capable of handling the downstream tasks such like producing a scalar reward or hate speech classification, enriched by its pretrained prior knowledge. Accordingly, the vertex embedding $F_v$ of each vertex is extracted from LLMs' last hidden layer to capture the textual features from inference conditioned on prompt and API description text. 
The graph adjacency matrix $A$ is an $N\times N$ matrix where nonzero entries indicate the presence of a connection, i.e., $A(x,y)\neq0$ if $x,y\in E(x,y)$. 
The vertex embeddings and the adjacency matrix together form the graph structure fed into the GNN.
In LLMEG, the  GNN encoder is implemented using two layers of graph convolution network (GCN) with the inputs as vertex textual features $F_v$, and the graph adjacency matrix $A$. The encoder outputs graph embeddings $e_g$ of size $N\times F'$, computed as:
\begin{equation}
    e_g=\phi(WAF_v)
\end{equation}
where $N$ is the number of vertexes, $F'$ is the dimension of the GCN layer, $W$ is a learned weighted matrix and $\phi$ is the LeakyReLU activation function. The graph embedding is concatenated with the word embeddings to form the feature embedding $e_f=[\eta e_g;e_w]$ as the input to the decoder, where $\eta$ is a trainable projection matrix that maps $e_g$ to the same latent space as $e_w$. The decoder $h(\cdot)$ consists of multi-layer perceptron (MLP) layers followed by a sigmoid activation function $\sigma$, producing the output:
\begin{equation}
    o = \sigma(h(e_f))
\end{equation}
The output, with size $N$, indicates whether each vertex should be selected for the given task. To train the LLMEG encoder-decoder, the cross-entropy loss 
\begin{equation}
    CE=-[vlog(\hat{v}) + (1-v)log(1-\hat{v})]
\end{equation}
 is used, where $v$ is the ground truth label indicating whether a vertex should be selected, and $\hat{v}$ is the predicted probability.

The result $o$ from the LLMEG result is used to construct an undirected graph that identifies the APIs required to complete the task. To determine the sequence of API executions, classical path-searching algorithms, such as Depth First Search (DFS), can be applied. However, DFS cannot select the appropriate starting point due to its lack of semantic understanding. To address this limitation, the LLM is utilized again to transform the undirected graph into a directed graph $G^*$, 
 incorporating semantic information from the subtasks. The system then executes the vertexes sequentially based on the order in the directed graph $G^*$.

\subsection{Ultrasound knowledge embedded and user intent understanding}
To adapt the model for understanding ultrasound tasks, a lightweight adaptation method, referred to as the LLaMA Adapter\cite{zhang2023llama}, is employed to efficiently fine-tune the original LLM into an instruction-following model specific to ultrasound-related tasks. The LLaMA Adapter achieves this by inserting prompted parameters into the top L layers of the N-layers LLaMA transformer model, enabling the model to acquire a new semantic understanding. In the USPilot framework, two adapters are trained: one based on ultrasound-related questions-answer pairs and the other tailored for ultrasound scanning tasks. 

\begin{figure}[t]
      \centering
    \includegraphics[width=0.78\linewidth]{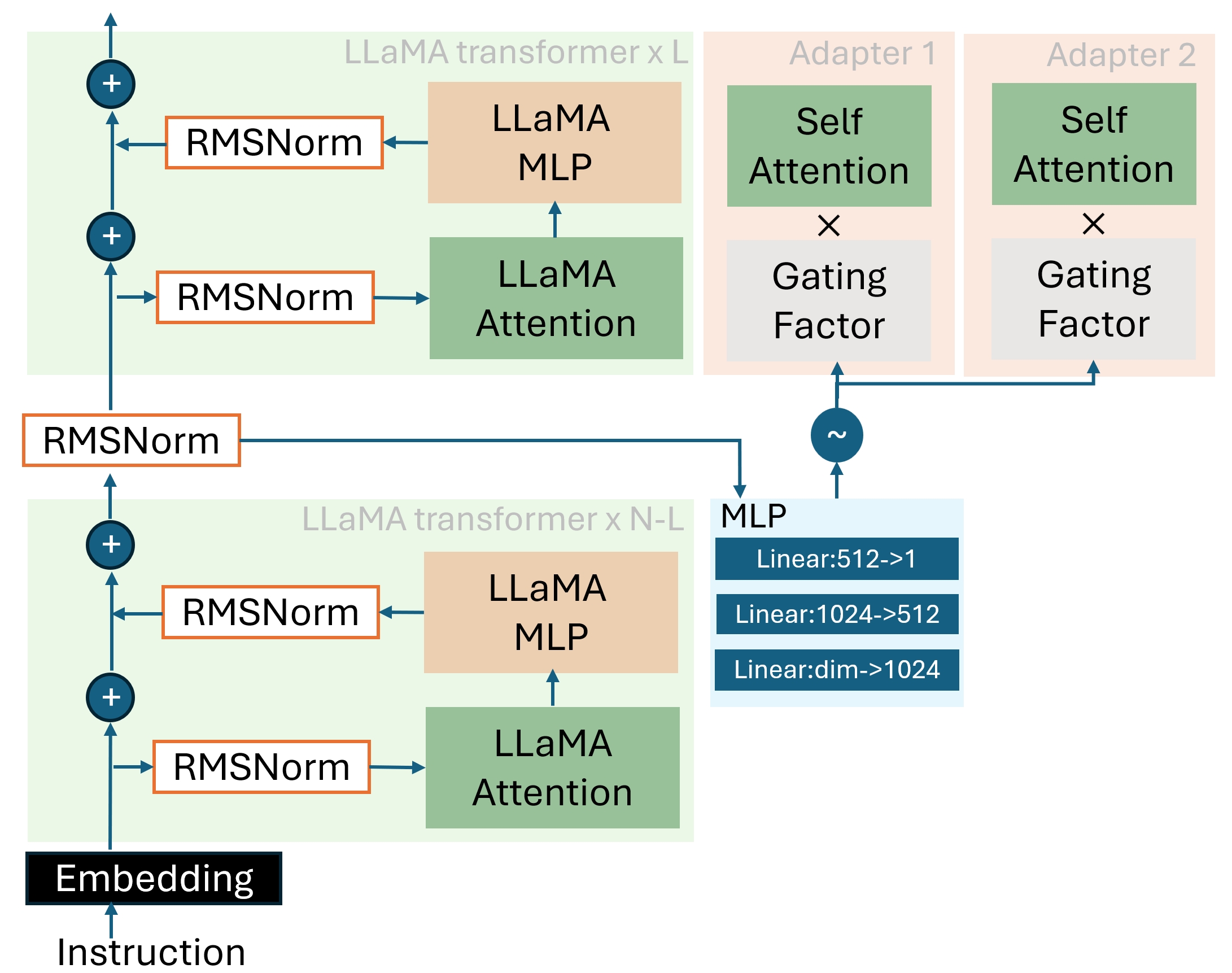}
      \caption{The dynamic routing mechanism switches the forward path among adapters according to the user’s instructions.}
      \label{mlp}
\end{figure}

Upon activating the LLaMA Adapter, the model experiences a reduced capability to handle general questions. As a comprehensive language model, retaining proficiency across a wide range of queries, including those outside the specific domain of robotic tasks is essential. However, the adapter structure may compromise the model's performance in general or medical question-answering scenarios while focusing on robotic tasks. A dynamic routing mechanism $r(\cdot)$ is integrated to address this limitation and restore versatility. The core component of this router is a three-layer MLP with a sigmoid activation gate. The complete architecture, including the dynamic router and adapters, is detailed in Fig. \ref{mlp}. The adaptation prompt for the l-th layer of the LLaMA adapter can be defined as the concatenated instruction knowledge learned prompt $P_l$ and word tokens $T_l$ as 
\begin{equation}
\begin{split}
[P_l;T_l]
\end{split}
\end{equation}
, where the learned prompt $P_l$ acts as a prefix turning. The dynamic router manages and routes tasks effectively based on semantic input. The learned adaptation prompt can be illustrated as 
\begin{equation}
    P_l=argmax(\sigma(r(T_{n-l})))*P_0+argmax(\sigma(r(T_{n-l})))*P_1
\end{equation}
, where $P_0$ and $P_1$ represent prompt vectors from different adapters. The training dataset is divided strategically between general question-answering tasks and specific robotic task instructions. This division enables the dynamic router to adapt efficiently between these two domains, balancing specialized and general capabilities. The router mitigates overfitting by appropriately routing tasks, ensuring the model retains general applicability while specializing in robotic tasks. Furthermore, the architecture demonstrates the potential for routing multiple questions or tasks using multiclass activation functions such as softmax. In this case, the routing weights can be expressed as: $P=onehot(argmax(\sigma(r(T_{n-l}))))*[P_0;P_1;P_2...P_n]^T$, where $\sigma$ refers to the softmax activation function.

\section{{Experiment}}
\label{sec:exp}
In this section, we present several experiments designed to assess the performance of the USPilot’s components. The LLMEG module was evaluated on a publicly available dataset compared to other methods. The semantic router and ultrasound knowledge adapter were also validated on a pre-collected dataset. Furthermore, the USPilot system was integrated with our robotic ultrasound scanning system, demonstrating its functionality in a physical environment.

\subsection{Datasets}
To demonstrate that LLMEG module learns instructed API selection and task planning, three public datasets from TaskBench\cite{shen2023taskbench}
were selected for evaluation, each with 500 test samples (1.5K in total). 
\textbf{Multimedia}: Contains 5.56K user-centric tasks such as file downloading and video editing. It is represented as a graph with 40 vertexes and 449 edges. \textbf{HuggingFace}: Covers various AI models covering 7.46K tasks across language, vision, audio, and video domains. Each task is treated as a tool for addressing specific problems, forming a graph with 23 vertexes and 225 edges. \textbf{Dailylife APIs}: Includes 4.32K tasks for daily services such as web search and shopping. It is represented as a graph with 40 vertexes and 1,560 edges. 

The semantic router and ultrasound knowledge adapter were trained and evaluated on two self-created datasets as followings. We also created an anonymous repository to share the dataset in the URL: \url{https://anonymous.4open.science/r/uspilot_data-0A0D/}

\textbf{Question Set}: Contains 13.3K entries, which comprise 10.5K question-answer(QA) pairs and 2.8K robotic instructions. The 10.5K QA pairs combine 9K samples from the public Alpaca dataset\cite{alpaca} 
and 1.5K in-house self-collected ultrasound-related QA pairs. Each entry includes four keys: ``instruction'' contains user question, ``input'' contains additional context, ``output'' is the expected generation result of the LLM is not used in any training steps and ``class'' is the labeled user intent. The following is an example:
\textit{\{``instruction'': ``Name 5 countries in the African continent.'',
``input'': ``'',
``output'': ``Nigeria, Egypt, South Africa, Ethiopia, and Ghana.'',
``class'': ``1''\}}.

\begin{figure}[t]
    \centering
    \includegraphics[width=1\linewidth]{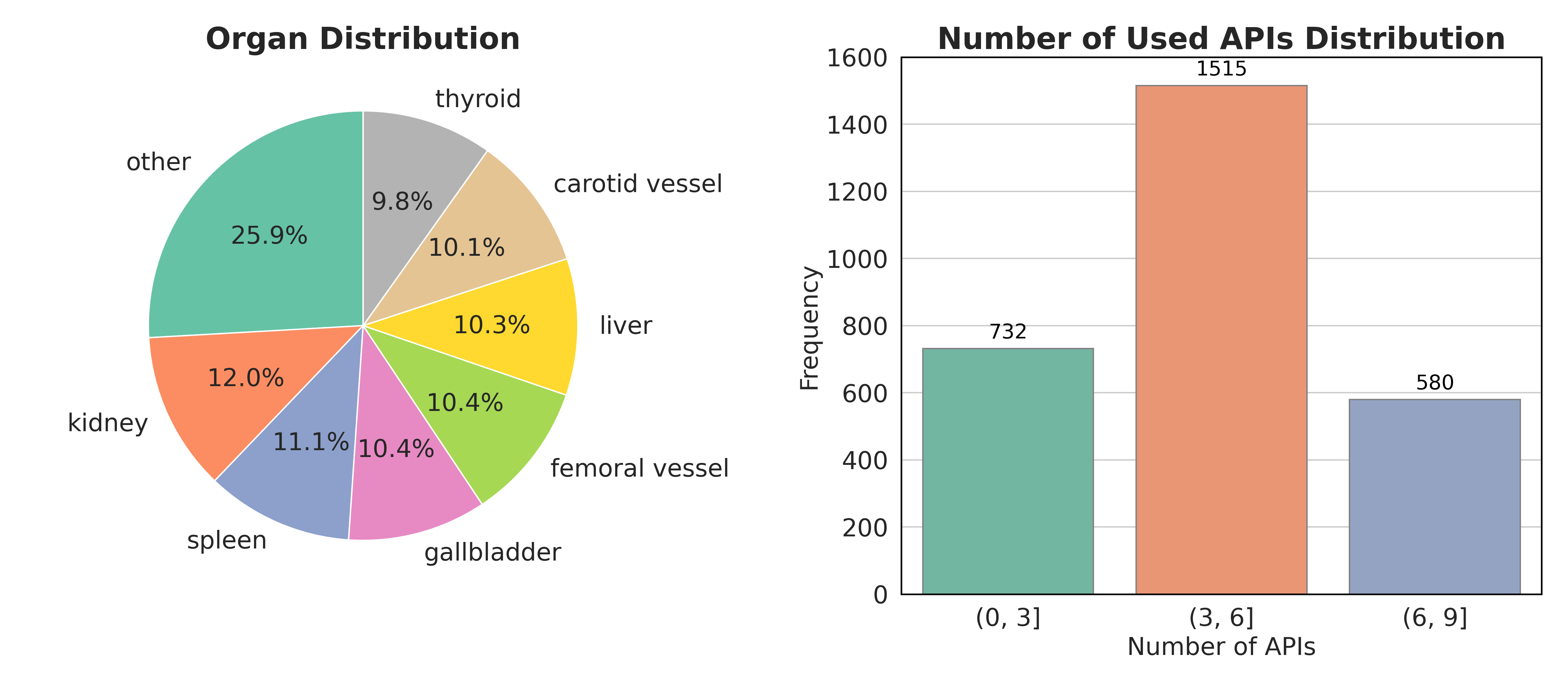}
    \caption{Categorical distribution on organs and numbers of used APIs}
    \label{re_cat}
\end{figure}

\textbf{Instruction Set}: Contains 2.8K fully collected in-house instructions. Describes operational procedures of the ultrasound robotic system, mainly focusing on seven body parts: carotid vessel, thyroid, liver, kidney, spleen, femoral artery, and gallbladder. The dataset comprises instructions and expected subtasks, in which the instructions have the patient’s direct command and symptom description, such as ``Scan the patient's liver'' and ``The patient has jaundice; check him with ultrasound.''. The complexity varies from requiring to use 1 API to 9 APIs as shown in Fig.\ref{re_cat}
. Each entry includes three fields: ``instruction'' is the input for the LLM, ``output'' is the expected generation result of it and ``API'' is the ground truth of the robot API excitation, but it is not used in the fine-tuning stage. Here is an example of the instruction set:
\textit{\{``instruction'': ``My left side feels tender to the touch.'',
``input'': ``'',
``output'': ``Step 1. Change the ultrasound probe to curvilinear probe. Step 2. Use camera
to detect the spleen area on upper left part of abdomen. Step 3. Execute robotic scanning.
Step 4. Segment spleen from real-time ultrasound image. Step 5. Publish report to
gastroenterology department. Step 6. Done'',
``API'': change\_probe(curvilinear),
detect\_organ(spleen),
execute\_robot(),
segment\_organ(spleen),
publish\_report(gastroenterology)\}}.

\vspace{-5pt}

\subsection{Comparison experiment for LLMEG}
\label{section_exp_llmeg}

The LLMEG was trained on 5K training samples and 500 test samples from each of Multimedia, HuggingFace and Dailylife APIs. For optimization, we used the AdamW optimizer with a learning rate of $1e-4$, betas of $(0.9, 0.95)$, a weight decay of 0.001, and a batch size of 64.
We employed 3 metrics to evaluate the methods' performance. \textbf{Accuracy} is the success rate of completing the given tasks. The \textbf{vertex F1 score} evaluates the labels assigned to individual vertexes, which can be indicated as 
\begin{equation}
    F1_{vertex}=\frac{2\times Precision_{vertex}\times Recall_{vertex}}{Precision_{vertex}+ Recall_{vertex}}
\end{equation}
, where the $Precision_{vertex}$ is the number of correctly identified vertexes among all predicted vertexes. $Recall_{vertex}$ is the number of correctly identified vertexes out of the total actual vertexes, representing correct API calls.
The \textbf{edge F1 score} measures the model's ability to predict valid or expected API call sequences, using the same F1 equation
\begin{equation}
     F1_{edge}=\frac{2\times Precision_{edge}\times Recall_{edge}}{Precision_{edge}+ Recall_{edge}}
\end{equation}
, where $Precision_{edge}$ and $Recall_{edge}$ follow the same definition as above, but applied on edges.

\begin{table*}[]\centering
\caption{Vertex F1 score, edge F1 score and accuracy across three datasets (Dailylife, Multimedia and Huggingface) for GNN4Task with GPT4-turbo, LLMEG (LLaMA3-8b) without subgraph, and LLMEG (LLaMA3-8b) with subgraph}
\begin{tabular}{cccccccccccc}
\hline
Method            & LLM        & Parameters & \multicolumn{3}{c}{Dailylife} & \multicolumn{3}{c}{Multimedia} & \multicolumn{3}{c}{Huggingface} \\
                  &            &            & Vertex F1   & Edge F1  & Acc    & Vertex F1   & Edge F1   & Acc    & Vertex F1    & Edge F1    & Acc   \\ \hline
GNN4TaskPlan      & GPT4-turbo & 8*222b     & 97.10     & 85.22    & 86.77      & 88.56     & 69.60     & 60.97      & 77.01      & 50.49      & 34.09     \\
Ours w/o subgraph & LLaMA3     & 8b         & 97.29     & 26.35    & 39.12  & 89.37     & 50.91     & 28.0   & 91.09      & 52.59      & 33.8     \\
Ours              & LLaMA3     & 8b         & 97.29     & 82.87    & 78.64  & 89.37     & 75.35     & 60.8   & 91.09      & 74.94      & 59.6    \\ \hline
\end{tabular}
\label{compare}
\end{table*}

\begin{figure*}
\centering
\begin{subfigure}{0.32\textwidth}
    \includegraphics[width=\textwidth]{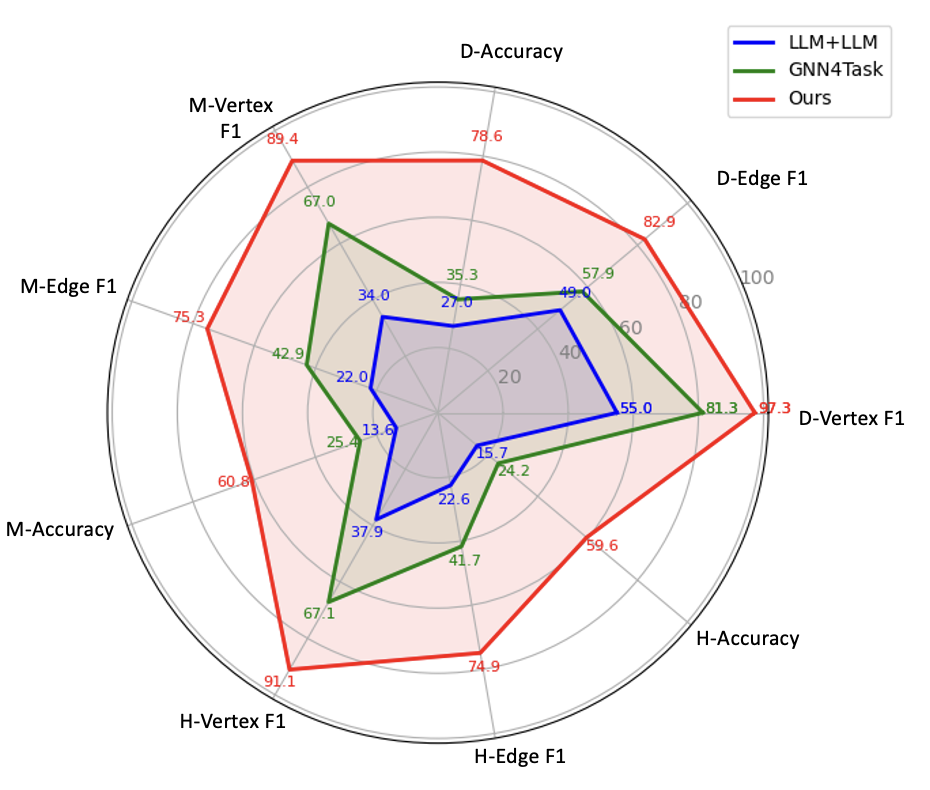}
    \caption{Results with LLaMA3-8b as base model}
    \label{fig:first}
\end{subfigure}
\hfill
\begin{subfigure}{0.32\textwidth}
    \includegraphics[width=\textwidth]{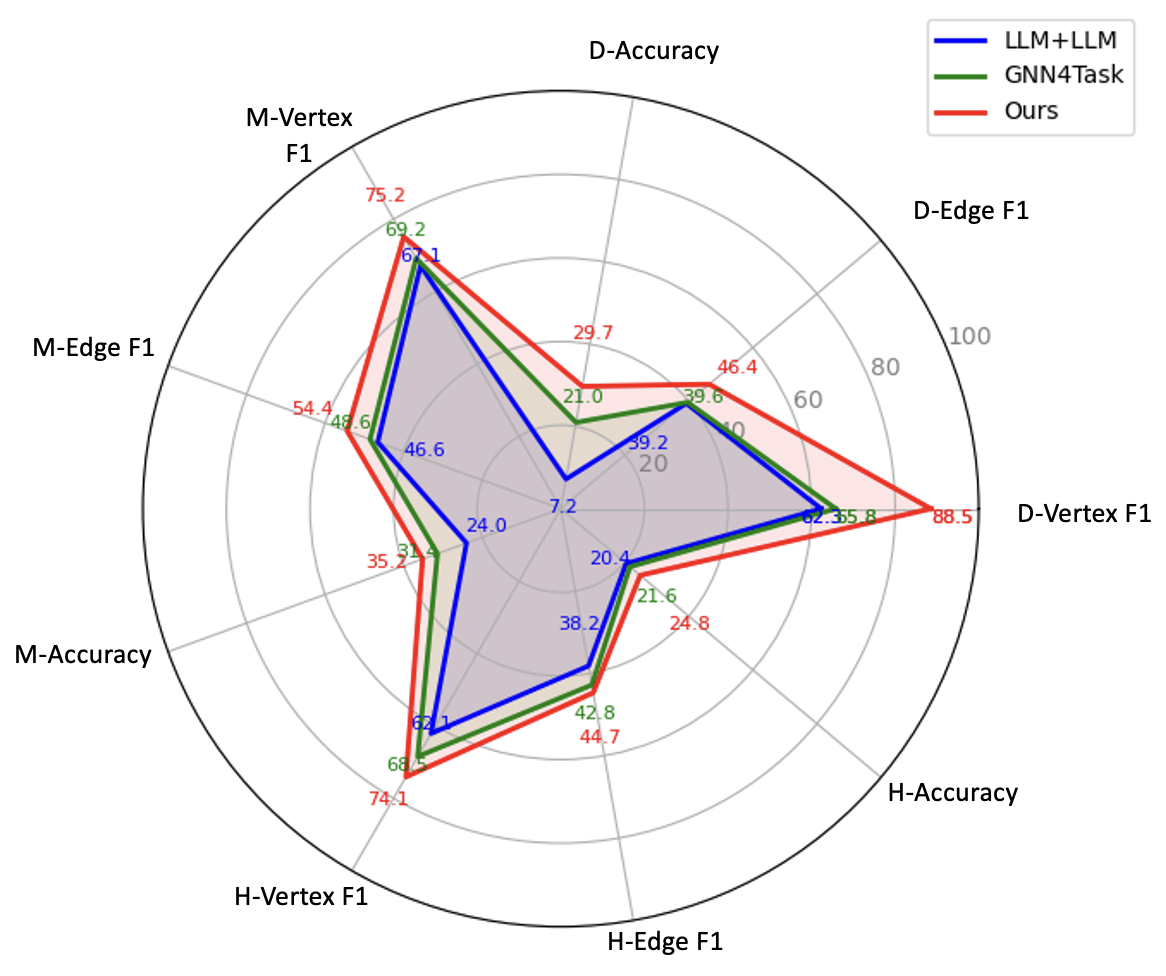}
    \caption{Results with Qwen2.5-7b as base model}
    \label{fig:second}
\end{subfigure}
\hfill
\begin{subfigure}{0.32\textwidth}
    \includegraphics[width=\textwidth]{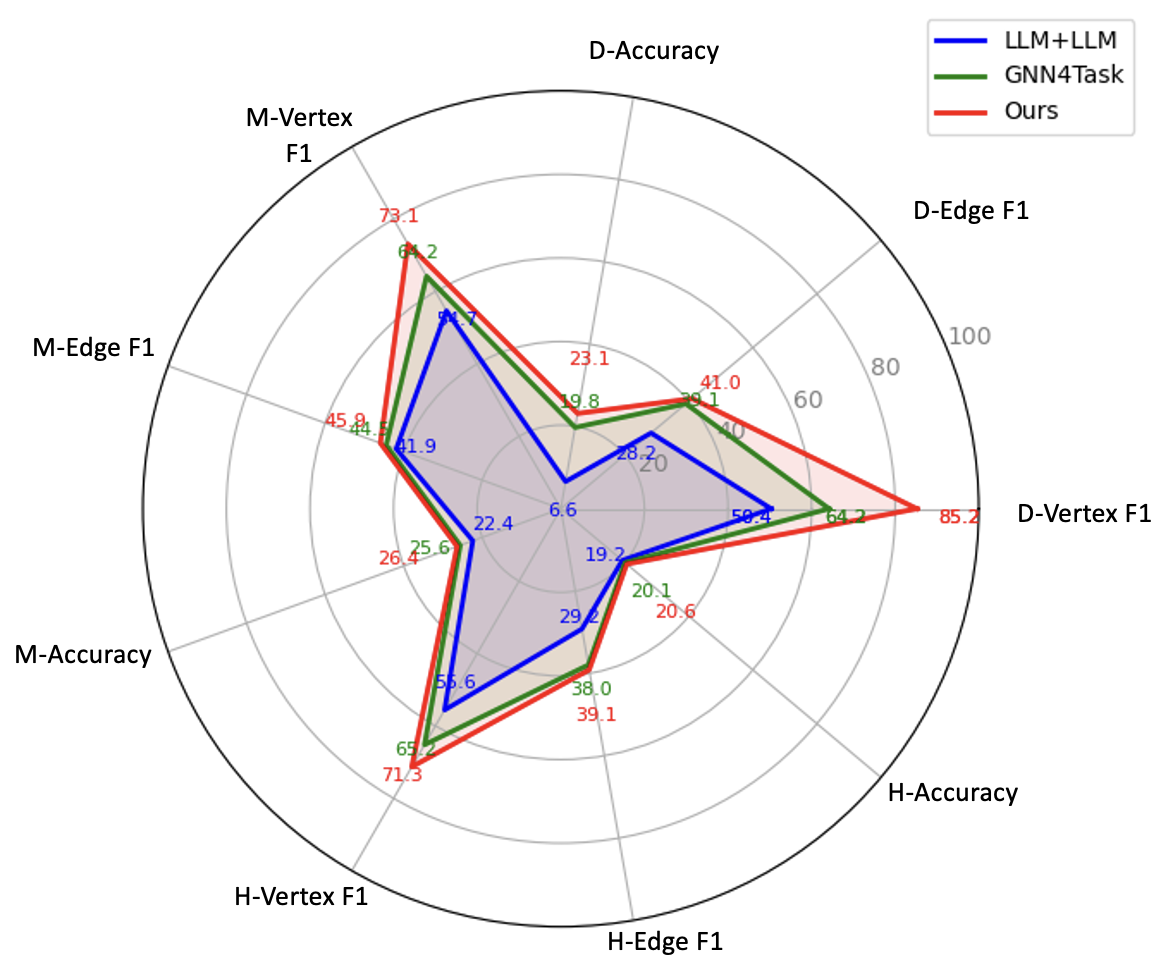}
    \caption{Results with Qwen2.5-3b as base model}
    \label{fig:third}
\end{subfigure}
\caption{Comparesion of LLM+LLM, GNN4Task, and our proposed method across Dailylife, Multimedia, and Huggingface datasets, where 'M-' indicates the Multimedia dataset, 'D-' indicates the Dailylife dataset and 'H-' represents the Huggingface dataset.}
\label{fig:multillm}
\end{figure*}

In the compression experiment, we employed two alternative methods:
\textbf{LLM+LLM}: This method involves collaboration between two LLMs. Initially, the text descriptions of all the tools and the user request are provided to the first LLM, which generates the necessary substeps to complete the task. Subsequently, the generated substeps are combined with the tool descriptions to form another prompt for the second LLM, which selects the tools to be utilized. The selected tools are considered as vertexes, while two adjacent vertexes can be regarded as edges. \textbf{GNN4Task}\cite{wu2024can}: The GNN4Task generates subtasks using LLMs and employs a small language model (e5-355M) to embed each subtask. A GNN processes each embedded subtask to identify vertexes.

We employed an 8-billion-parameter LLaMA3 model in our proposed method LLMEG and compared it with GNN4Task, which is equipped with one of the powerful LLMs, GPT4-tubo, a mixture of eight 222 billion parameters experts. As shown in Table 
\ref{compare}
, the proposed LLMEG achieves superior performance across all three public datasets. On the Dailylife dataset, both methods perform well due to the inherent ability of LLMs to understand daily service descriptions. On the Multimedia dataset, our method slightly outperforms GNN4Task. On the more professional HuggingFace dataset, it significantly enhances vertex selection and edge prediction. We also compared our method with a variant using DFS instead of the LLM-generated subgraph. Subgraph planning achieves improvements on both edge F1 score and total accuracy, especially on Dailylife dataset. This highlights the importance of semantic understanding in generating correct API sequences, even with accurate vertex selection. The largest gain on Dailylife, which contains 39 edges per vertex, showing that semantic matters more as graph complexity increases.

After employing the LLMEG, we conducted experiments using different LLMs with varying parameter sizes to assess planning capabilities. As shown in Fig. \ref{fig:first}, our proposed method outperforms the original LLM-based method and GNN4Task in both vertex and edge F1 scores by over 40\% across all datasets when using LLaMA3-8b and its embeddings. For other LLMs who cannot generate original sentence embeddings, LLMEG abstracts word embedding with 137M parameters small language model (nomic-embed-textv1.5). Our proposed method achieves better performance across all metrics for a similar size Qwen2.5-7b. However, when integrated with smaller LLMs shown in Fig. \ref{fig:third}, our LLMEG can conduct a high value only in vertex F1 score but struggles with edge F1 scores and accuracy. This is due to smaller LLMs' limited semantic understanding capabilities, which are insufficient for effective subgraph replanning.

\subsection{User intent understanding and ultrasound knowledge embedded}
\label{sec:user_intent}
The user intent prediction MLP router was trained with 1.2K data randomly split from the question set with an equal ratio of 1:1:1 for 100 epochs. We used the AdamW optimizer with a learning rate of $1e-6$ and a batch size of 1. We used a sigmoid as the activation function to classify the user inputs as QA inquiries or ultrasound robot commands. We also deployed a direct fine-tuning method by mixing the QA and robotic execution data to fine-tune the LLM. The fine-tune is on the same 1.2K train set as the MLP with 10 epochs. Adamw optimizer was used with a learning rate of 1e-5, and a batch size of 1. As shown in Table \ref{tcc}, the model was evaluated on the rest of the question set. The results demonstrate that the MLP layer can reach a 95\% classification accuracy and the direct fine-tune one can only reach 91\% accuracy. 
\begin{table}[]\centering
\caption{Accuracy of user intent classification by USPilot}
\begin{tabular}{ccccc}
\hline
         & \multicolumn{2}{c}{Router} & \multicolumn{2}{c}{Finetune} \\
         & precision      & recall     & precision       & recall      \\ \hline
QA   & 0.9           & 0.99       & 0.97           & 0.97        \\
Robot      & 1.0           & 0.98       & 0.87           & 0.84        \\
Accuracy & \multicolumn{2}{c}{0.98}   & \multicolumn{2}{c}{0.91}     \\ \hline
\end{tabular}
\label{tcc}
\end{table}

\begin{table}[]\centering
\caption{Success rate of the fine-tuned USPilot adapter in direct commanding and symptom description across seven seen body parts, four operation situations, and two unseen body parts}
\begin{tabular}{ccc}
\hline
               & Direct Command  & Symptom Description \\ \hline
Seen           &                 &                     \\ \cline{1-1}
Carotid Artery & 95\%            & 96\%                \\
Thyroid        & 100\%           & 85\%                \\
Liver          & 100\%           & 85\%                \\
Kidney         & 95\%            & 75\%                \\
Spleen         & 95\%            & 85                  \\
Femoral Artery & 90\%             & 75\%                \\
Gallbladder    & 85\%            & 80\%                \\
\textbf{Total} & \textbf{95\%}   & \textbf{78.5\%}     \\ \cline{1-1}
Unseen         &                 &                     \\ \cline{1-1}
Bladder        & 37.5\%          & 15.5\%              \\
Pancreas       & 94.7\%          & 47.3\%              \\
\textbf{Total} & \textbf{71.4\%} & \textbf{25.7\%}     \\ \hline
Single API        &                 &                     \\ \cline{1-1}
Interrupt      & 100\%           & -                   \\
Continue       & 100\%           & -                   \\
Increase force & 100\%            & -                   \\
Decrease force & 95\%           & -                   \\
\textbf{Total} & \textbf{99\%}   & -                   \\ \hline
\end{tabular}
\label{sr}
\end{table}
In addition, the ultrasound robotic adapter was fine-tuned on the whole instruction set for 100 epochs. The training set and test set are split with a ratio of 9:1. AdamW optimizer was used with a learning rate of $1e-5$, and a batch size of 1. Furthermore, the bladder and pancreas, which did not appear in the training set, were considered as unseen scenes for evaluation to demonstrate the LLM's ability to transform the knowledge to similar scenarios.
The Success Rate (SR) is the fraction of successfully completed instructions. As Table \ref{sr} indicates, USPilot achieved a 99\% success rate for single API executions. The fine-tuned adapter can comprehend the instructions, achieving a 95\% success rate for direct commands and 78.5\% for symptom-based descriptions involving seen body parts. For unseen body parts (bladder and pancreas), USPilot leveraged the LLM’s generalization ability to achieve a 71.4\% success rate for direct commands. However, it achieved only a 25.7\% success rate for symptom-based descriptions due to the absence of specific knowledge about these body parts in the training data. Instead of effectively generalizing to entirely unseen body parts, it exhibited knowledge transfer between related body parts, such as from the gallbladder to the pancreas or from the kidney to the bladder, which resulted in lower success rates.

\subsection{{USPilot in the real world ultrasound scenario}}
\label{sec:realworld}
 The system comprises an ultrasound robotic system, a 3D camera system, and a high-performance computing server. To ensure patient safety, the ultrasound robotic system performs pre-defined trajectory scanning with admittance control. It adjusts the probe’s orientation based on the contact point to achieve an optimal contact angle with the skin, ensuring the acquisition of clear ultrasound images\cite{cao2023ultra}. To determine the scanning trajectory and cover various body parts, the system integrates a patient anatomy avatar reconstruction model\cite{zhou2024inverse}, which provides the positions of key anatomical structures such as the carotid artery, thyroid, and liver. USPilot receives instructions from the sonographer, who can pose medical questions or command the robotic system to perform ultrasound scans.

\begin{figure}
\centering
\begin{subfigure}{0.38\textwidth}
    \includegraphics[width=\linewidth]{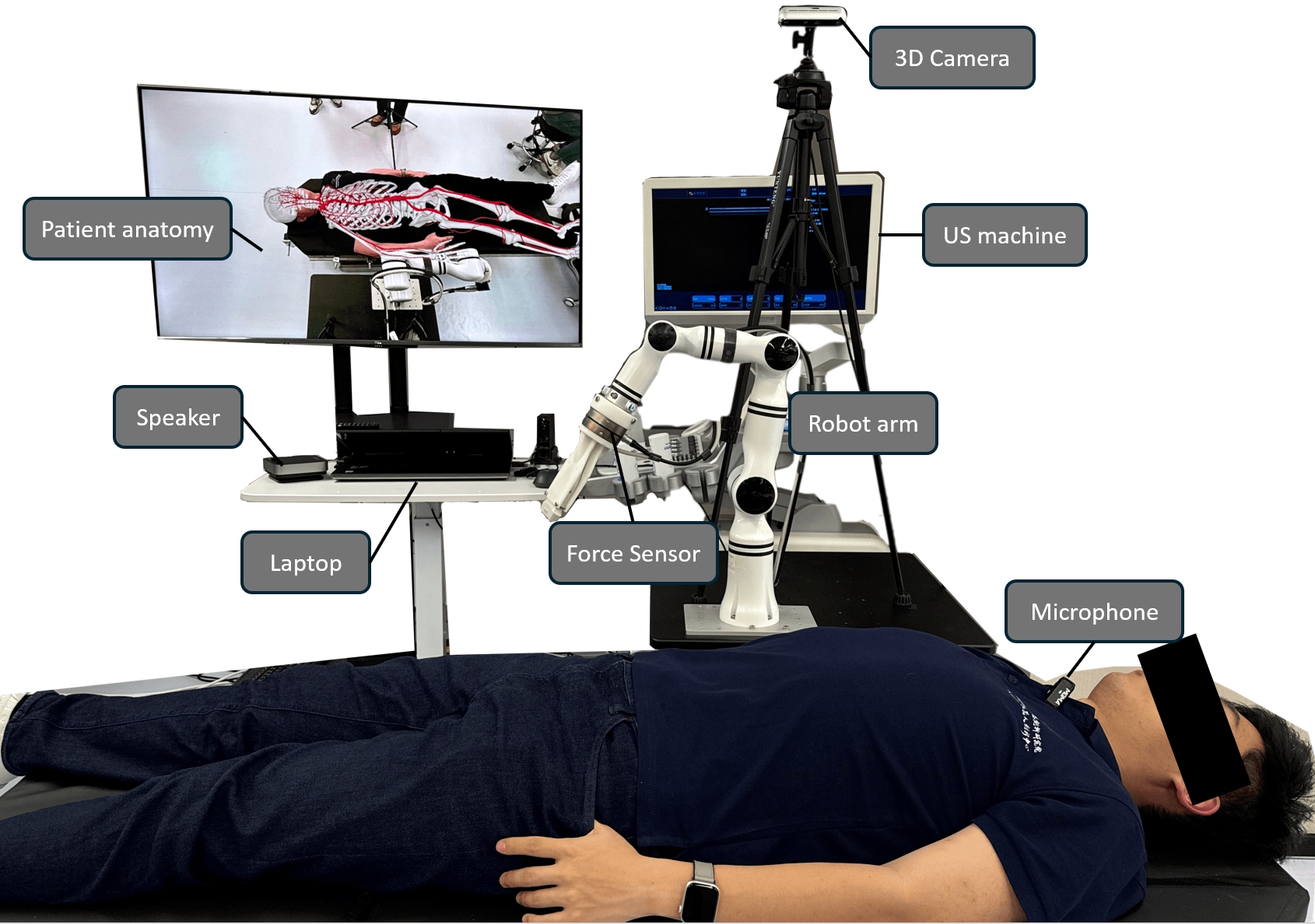}
      \caption{The Physical setup of USPilot.}
      \label{fig:setup}
\end{subfigure}
\hfill
\begin{subfigure}{0.42\textwidth}
    \includegraphics[width=\linewidth]{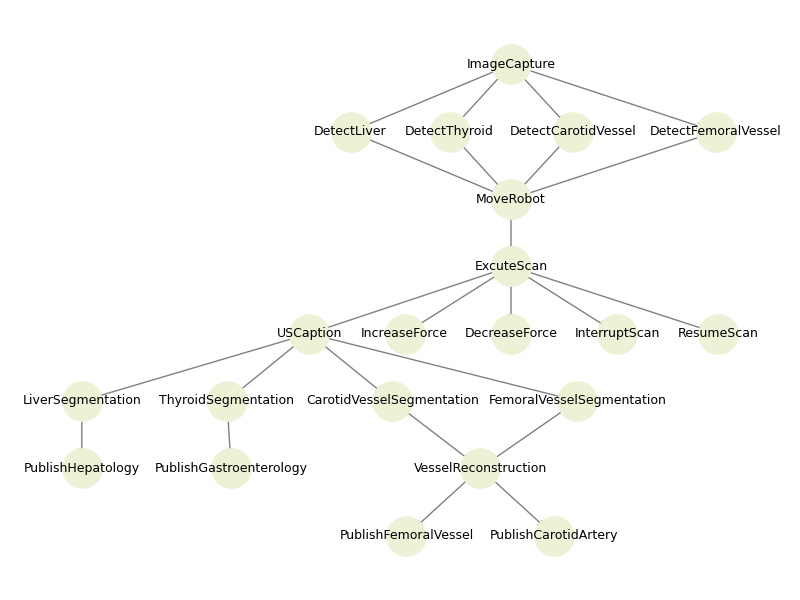}
      \caption{The graph structure of USPilot.}
      \label{fig:uspg}
\end{subfigure}
  
\caption{USPilot in a Real World Ultrasound Scenario}
\label{fig:fullset}
\end{figure}

As shown in Fig. \ref{fig:setup}, the physical setup consists of an ultrasound system (Pioneer H20, Angell, China) equipped with a linear probe and a convex probe mounted on a 6-axis collaborative robotic arm (RM65, Realman, China). A probe fast assemble machine is positioned adjacent to the robot system. A 3D camera (Realsense D455, Intel, US) mounted on the robot platform provides anatomical structure positioning and patient orientation information, serving as part of the system’s environmental feedback. With the patient anatomy structure, the system determines the initial scanning position for different targets. A 6-axis force sensor (M3815D, SRI, China) is integrated into the robot's end-effector to ensure patient safety and provide orientation feedback for the system. There are 21 APIs and 24 edges defined in the USPilot system, as depicted in the graph structure in Fig. \ref{fig:uspg}. The system employs a wake-word listener for human-machine interaction to ensure accessibility and ease of use. The user can activate the system by calling ``Doctor" or ``Hey, Doctor". Upon activation, the user's speech is processed and transcribed into text to USPilot. The system can then convert the generated text or pre-defined API text into audio, which is played back to the user.

The computational requirements of the proposed system are influenced by the inference latency, which largely depends on the content and length of the user request. For non-quantized models operating in FP16 precision, LLaMA3-8b requires approximately 20 GB of GPU memory. In our real-world ultrasound scenario experiment, the robot was controlled by a local PC, while the model ran on a local server equipped with a 24 GB Nvidia 4090 GPU. The total latency consists of three main components: 1) The router and request decomposition process requires 0.029 seconds per inference token. On average, this results in a latency of 0.6 seconds per single-step request on our dataset. 2) Using LLaMA3-8b as the backbone, the graph generation process takes an average of 0.13 seconds to produce the final result. 3) Communication between the PC and the local server is handled via Redis, introducing an average latency of 0.47 seconds per interaction.

 \begin{table*}[]\centering
 \caption{Results from the real-world ultrasound scan with USPilot}
\begin{tabular}{llll}
\hline
Type           & Num & Instruction example                                                                                  & SR    \\ \hline
Carotid vessel & 10  & I got dizzy recently, the doctor said my carotid vessel may be blocked.                                & 10/10 \\
Liver          & 10  & My physical examination report shows that my transaminase is elevated. Please help me check my body. & 10/10 \\
Force adjust   & 5   & You pressed me a little pain.                                                                      & 3/5   \\
Interrupt       & 5   & I feel not good.                                                                                      & 5/5   \\ \hline
\end{tabular}
\end{table*}

\begin{figure}[t]
    \centering
    \includegraphics[width=0.94\linewidth]{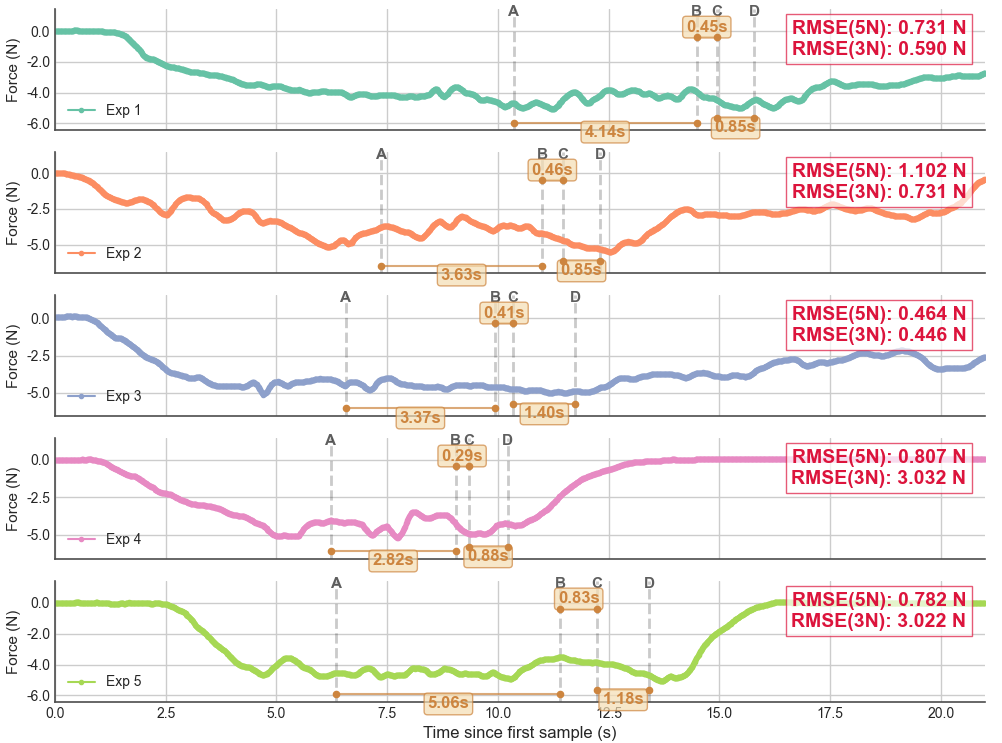}
    \caption{The USPilot force result with instructions to reduce force}
    \label{force}
\end{figure}

 In the real-world scenes, the USPilot was evaluated on two body targets with 10 testing rounds per target and 5 additional rounds for single API commands in Table \ref{sr}. The robot was required to perform automatic scanning based on varying instructions for each target. We used the success rate as the evaluation metric, defined as the proportion of test rounds in which the robots successfully completed all assigned tasks. In the scanning task, the robot can successfully locate the commanded target and perform the scan with a 100\% success rate. For the force adjustment task, USPilot achieved a success rate of only 60\%, as shown in Fig. \ref{force} , which records the contact force during the experiment. The two failure cases are due to a misunderstanding of the word ``painful'' of the LLM to interrupt the scan. The time costs were analyzed across three stages: A-B represents the user's speaking period, B-C corresponds to the speech-to-text transcription time, which ranged from 0.29s to 0.83s, and C-D denotes the LLM and graph processing time, spanning approximately 0.8s to 1.4s. The initial contact force was set to 5 N, with a root mean square error (RMSE) ranging from 0.464 N to 1.102 N. After processing the user's instructions, the expected force was reduced to 3 N, resulting in an RMSE of 0.446 N to 0.731 N across experiments 1 through 3.

\section{Discussion \& Future work}
The results in Table \ref{compare} demonstrate that the proposed LLMEG method with LLaMA3-8b surpasses GNN4TaskPlan with GPT-4turbo-8*222b. As shown in Fig. \ref{fig:multillm}, reducing word embedding size significantly degrades the performance of the proposed method, preventing it from matching the metrics of the LLaMA3-8b model. Moreover, smaller LLM sizes lead to substantial performance declines, highlighting the strong influence of LLM-generated steps on the GNN classification model. The experiments also reveal that, even with optimal API selection, the order of API calls is heavily dependent on the LLM's planning capability.  As discussed in Section \ref{sec:user_intent}, the proposed USPilot system offers potential for an autonomous robotic ultrasound platform by effectively distinguishing user inputs as either questions or robot-executable commands. While the fine-tuned adapter performs well in executing direct commands and interpreting symptom descriptions, the dataset remains too limited to ensure reliable performance in medical scenarios. Expanding the dataset with more diverse language instructions will be critical for further fine-tuning. In section \ref{sec:realworld}, we demonstrated the USPilot with the physical setup to perform real ultrasound scans on the human body. 

To improve the stability and accuracy, the proposed LLMEG still relies on strong prior knowledge of the robotic system, which limits adaptability. A more generalized approach of using two collaborating LLMs was also evaluated in Section \ref{section_exp_llmeg}
, which requires no extra prior knowledge but performed the worst as shown in Fig. \ref{fig:multillm} 
. This sort of trade-off is necessary under our fixed self-built robotic system and protocol-driven nature of ultrasound scanning, where high generalization is typically unnecessary. Learning on priors is reasonable for stability and accuracy, but exploring to reduce training cost remains important.

The proposed method focuses on planning ultrasound robotic procedures, introducing a novel human-robot interaction framework for an unmanned ultrasound system. However, several areas warrant further exploration. Firstly, while the system effectively serves as a planner, arranging robotic tool sequences based on diverse user inputs and acting as a ``virtual doctor'' to answer questions via a semantic router, it heavily relies on predefined APIs. These APIs dictate the rule-based control logic, limiting the system’s adaptability and scalability. A key challenge during development was designing an API structure that balanced diverse user instructions with compatibility to rule-based mechanisms. With advancements in multimodal vision-language-action models in collaborative robotics, future work should integrate real-time probe control into the framework. Such models could process not only language inputs but also image and force data, enabling adaptive, context-aware manipulation during ultrasound procedures. Secondly, the graph-based structure in the proposed method uses undirected graph features due to the nature of the GNN. This limits its ability to handle scenarios requiring cyclic graph structures, which are important for dynamic replanning and learning from errors. Incorporating directed graph features or alternative learning mechanisms will be essential to enhance the system’s robustness in managing unexpected scenarios or user modifications. While the current method provides a front-end for an automated sonographer, developing a more flexible and intelligent back-end remains a critical challenge for future research.

\section{Conclusion}

This paper introduces USPilot, an embodied framework for an autonomous ultrasound robotic system powered by an LLM-enhanced GNN. 
Through evaluations of LLMEG, semantic routers on datasets, and robot performance in real-world scenarios, USPilot shows stable performance in ultrasound robotic tasks and enhanced capabilities in managing ultrasound-related QA inquiries. By reimplementing existing methods across datasets derived from TaskBench and modifying the base model, the system exhibits a degree of knowledge transfer and adaptability to different domains. However, the size of the base model parameters remains a critical factor influencing performance, and the system lacks the ability to learn from past errors. Addressing these limitations represents a promising direction for future research on autonomous ultrasound robotic systems.


\bibliographystyle{ieeetr}
\bibliography{bibliography}

\end{document}